# Early Diagnostic Prediction of Covid-19 using Gradient-Boosting Machine Model

Satvik Tripath
*College of Computing & Informatics*
*Drexel Universiuty*
Philadelphia, USA
St3263@drexel.edu

*Abstract*—With the huge spike in the COVID-19 cases across the globe and reverse transcriptase-polymerase chain reaction (RT-PCR) test remains a key component for rapid and accurate detection of severe acute respiratory syndrome coronavirus 2 (SARS-CoV-2). In recent months there has been an acute shortage of medical supplies in developing countries, especially a lack of RT-PCR testing resulting in delayed patient care and high infection rates. We present a gradient-boosting machine model that predicts the diagnostics result of SARS-CoV-2 in an RT-PCR test by utilizing eight binary features. We used the publicly available nationwide dataset released by the Israeli Ministry of Health.

## I. INTRODUCTION

The Covid-19 Pandemic has continued to threaten the humanity and challenge the medical facilities and institutions. The demands for hospital beds and crucial lack of medical equipment are rising sharply, with several health professionals getting infected. Therefore, a quick, efficient clinical decision as well as optimized usage of resources is very necessary. The reverse transcriptase-polymerase chain reaction (RT-PCR) test remains a key component for rapid and accurate detection of SARS-CoV-2 infection [1].

There have been various diagnostic and prognostics model built that uses several features such as computer tomography (CT) scans, MRI, histopathology reports, clinical symptoms, and laboratory tests to help clinicians and medical staff in triaging patients. Moreover, helping in estimating growth and pattern of infection. Thus machine learning model can not only help in prediction of COVID-19 but also in preventing its spread [2].

## II. METHOD

We developed a state-of-the-art gradient-boosting machine with decision tree base learners architecture [3] for predicting tabular data. We utilized the LightGBM Python package along with the gradient boosting predictor for best classification on tabular data [4]. We trained and tested our model using the Israeli Ministry of Health's publicly released data generated by the SARS-CoV-2 results of individuals through RT-PCR assay of a nasopharyngeal swab. Both negative and positive cases were confirmed using the same test [5]. In addition to the test results, the dataset contained eight binary features as shown in Table 1 [6].

The training dataset contained records from 51,831 individuals (out of which 4769 were COVID positive cases) collected from March 22nd to March 31st, 2020. The test dataset contained test results of 47,401 individuals (out of which 3624were COVID positive cases), recorded from April 1st to April 7th, 2020 [6].

Area Under the Receiver Characteristic Operator (auROC) curves were used as the basis of evaluation to score the model's performance on the test dataset. Also, across varied thresholds, the area under the precision-recall (auPRC) curve was drawn. For a deeper evaluation of the model, metrics such as overall accuracy, confidence intervals (CI), specificity, specificity, positive predictive values (PPV), negative predictive values (NPV), false-negative rate, false-positive rate, and false discovery rate were calculated for every threshold from all of the auROC curves.

TABLE I. FEATURES AND CHARACTERISTICS OF THE DATASET USED IN THE RESEARCH

| Feature | Total | | Covid-19 Negatives | | Covid-19 Positives | |
|---|---|---|---|---|---|---|
| | n | % | n | % | n | % |
| *(1) Sex* | | | | | | |
| Male | 50,350 | 50.74 | 45,545 | 50.1 | 4805 | 57.2 |
| Female | 48,882 | 49.26 | 45,294 | 49.8 | 3588 | 42.7 |
| *(2) Age 60+* | | | | | | |
| True | 15,279 | 15.4 | 13,619 | 14.9 | 1660 | 19.7 |
| False | 83,953 | 84.6 | 77,220 | 85 | 6733 | 80.2 |
| *(3) Cough* | | | | | | |
| True | 14,768 | 14.88 | 10,715 | 11.8 | 4053 | 48.2 |
| False | 84,223 | 84.87 | 79,909 | 87.9 | 4314 | 51.4 |
| *(4) Fever* | | | | | | |
| True | 8122 | 8.18 | 4387 | 4.83 | 3735 | 44.5 |
| False | 90,868 | 91.5 | 86,237 | 94.9 | 4631 | 55.41 |
| *(5) Sore throat* | | | | | | |
| True | 8122 | 8.18 | 4387 | 4.83 | 3735 | 44.5 |
| False | 90,868 | 91.5 | 86,237 | 94.9 | 4631 | 55.1 |
| *(6) Shortness of breath* | | | | | | |
| True | 930 | 0.94 | 71 | 0.08 | 859 | 10.2 |
| False | 95,405 | 96.14 | 88,084 | 96.9 | 7321 | 87.2 |
| *(7) Headache* | | | | | | |
| True | 1799 | 1.81 | 68 | 0.07 | 1731 | 20.6 |
| False | 95,536 | 95.27 | 88,087 | 96.9 | 6449 | 76.8 |
| *(8) Confirmed contact with COVID-19 positive individual* | | | | | | |
| True | 5507 | 5.55 | 1455 | 1.6 | 4052 | 48.2 |
| False | 93,725 | 94.45 | 89,384 | 98.4 | 4341 | 51.8 |

## III. RESULTS

SHapley Additive exPlanations (SHAP) beeswarm plot were drawn to summarize and rank the most features as per their importance in the classifier's prediction (Fig.1). These values are generally useful for complex machine learning models such as Random forest classifiers, Gradient-boosting machines [7], or Artificial Neural Networks [8]. Features plotted on the y-axis are organized as per their Mean Absolute SHAP values. Each point represents an individual from the study. The positions of points on the x-axis describe the impact of the feature on the model's prediction. Color correlation is used to visualize feature value. So, we can infer that Cough, Fever, and Contact with confirmed are the key features in making predictions, also describing the importance of social distancing [9].

The model achieved an auROC of 0.90 with 95% CI: 0.892–0.905 obtained by bootstrapping (Fig. 2a). Through the predictions made by the classifier on the test dataset, the optimum working points would be 79.18% or 71.98% specificity and 87.30% or 85.76% sensitivity. The model also obtained an 0.66 auPRC with 95% CI: 0.647–0.678 (Fig. 2b).

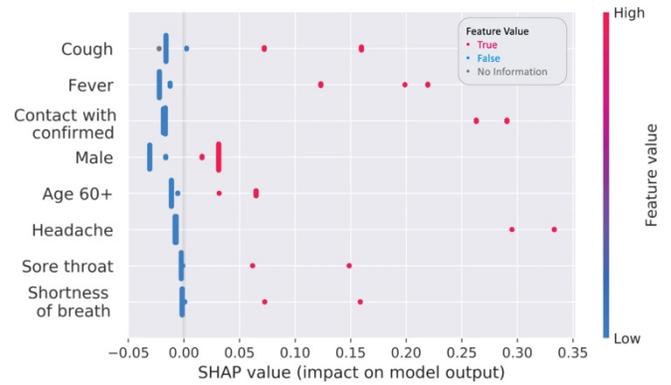

Fig. 1. Eight binary features. SHapley Additive exPlanations (SHAP) beeswarm plot

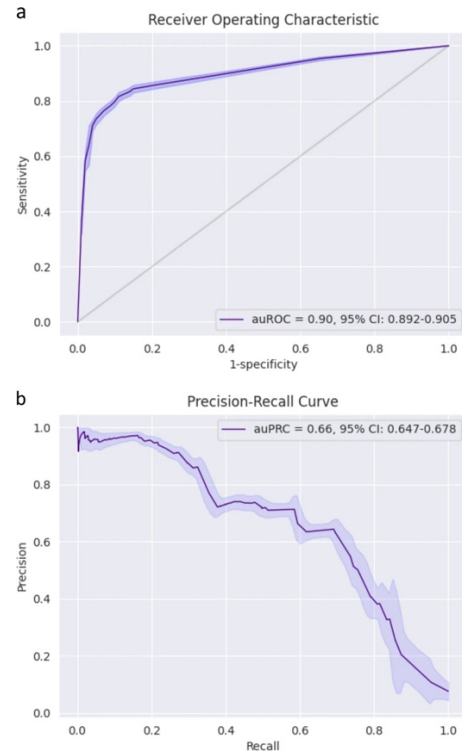

Fig. 2. . Model metrics evaluation. a) ROC curve of the model on the test dataset and shaded band around the curve representing confidence intervals (CI) obtained by bootstrapping. b) Area under the precision-recall (auPRC)curve for various thresholds.

## IV. DISCUSSION

The proposed methodologies and research work is indeed not free of shortcomings [10]. The dataset released by the Israeli Ministry of Health did have restrictions, biases, and a few important features were completely absent. For instance, for patients having had confirmed contact with a COVID-19 positive individual, more information such as duration of contact, location of contact (outdoor/indoor) was not available6. Moreover, some major clinical symptoms of COVID-19 infection, like lack of taste and smell, which could have helped in a much more accurate classification, were completely absent from the dataset5. Also, one of the limitations of the dataset, the patients who tested positive for COVID-19 had much more comprehensive symptoms reporting and were epidemiologically validated. The proportion of who were COVID-19 positive to patients who turned out positive in each feature. We identified that features like headache (96.2%), sore throat (92.3%), and shortness of breath (92.4%) have biased reporting. Whereas, cough (27.4%) and fever (45.9%) remained balanced features [11]. Therefore, the patients who tested negative for COVID-19 may have under-reported and mislabelled symptoms.

As a probable solution to simulate reduced biases in the model's validation metrics and test dataset, we randomly picked negative reports of patients across all eight symptoms and removed their negative values [12]. Our proposed model showed very promising results and achieved a high auROC score and accuracy on these three varied simulated datasets, thus establishing strong confidence in the effectiveness of the model (Fig. 3).

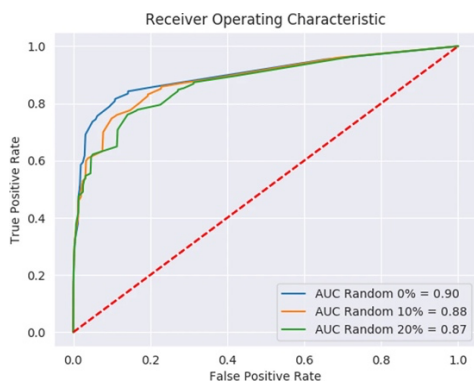

Fig. 3. Eight binary features. SHapley Additive exPlanations (SHAP) beeswarm plot

## V. Conclusin

We conclude that our proposed methodology can be implemented in a medical setting to provide an unbiased detection of covid-19 and to prioritize testing on the basis of the severity of the patient. We also investigated that the national data reported by the Israeli Ministry of Health contains some limitations and biases but with minor pre-processing and feature selection can be used to train models for the COVID-19 healthcare framework. In addition, the model presented may also help improve the existing health system response to any future pandemic waves of COVID-19 or any other acute respiratory disorder.

## References

[1] Hu, B., Guo, H., Zhou, P. & Shi, Z.-L. Characteristics of sars-cov-2 and covid-19. *Nat. Rev. Microbiol.* 1–14 (2020).

[2] Tripathi, S. Artificial intelligence: A brief review. *Anal. Futur. Appl. AI, Sensors, Robotics Soc.* 1–16 (2021).

[3] Chen, T. & Guestrin, C. Xgboost. *Proc. 22nd ACM SIGKDD Int. Conf. on Knowl. Discov. Data Min.* DOI: 10.1145/ 2939672.2939785 (2016).

[4] Lundberg, S. M., Lee, S.-I. & Guyon, I. Advances in neural information processing systems. *New York: Curran Assoc.* (2017).

[5] Covid-19-government data information. https://data.gov.il/dataset/covid-19/resource/ 3f5c975e-7196-454b-8c5b-ef85881f78db/download/-readme.pdf (2020).

[6] The novel coronavirus israel ministry of health. https://govextra.gov.il/ministry-of-health/corona/corona-virus-en/ (2020).

[7] Hastie, T., Tibshirani, R. & Friedman, J. *The elements of statistical learning: data mining, inference, and prediction* (Springer Science & Business Media, 2009).

[8] Lundberg, S. M. *et al.* Explainable machine-learning predictions for the prevention of hypoxaemia during surgery. *Nat. biomedical engineering* 2, 749–760 (2018).

[9] Liu, Y., Gayle, A. A., Wilder-Smith, A. & Rocklöv, J. The reproductive number of covid-19 is higher compared to sars coronavirus. *J. travel medicine* (2020).

[10] Morley, J. *et al.* The ethics of ai in health care: A mapping review. *Soc. Sci. & Medicine* 113172 (2020).

[11] Covid-19-government data. https://data.gov.il/dataset/covid-19 (2020).

[12] Rigby, M. J. Ethical dimensions of using artificial intelligence in health care. *AMA J. Ethics* 21, 121–124 (2019).